# Building subject-aligned comparable corpora and mining it for truly parallel sentence pairs.

Krzysztof Wołk, Krzysztof Marasek
Polish Japanese Institute of Information Technology, Warsaw, Poland
kwolk@pjwstk.edu.pl

**Abstract**

Parallel sentences are a relatively scarce but extremely useful resource for many applications including cross-lingual retrieval and statistical machine translation. This research explores our methodology for mining such data from previously obtained comparable corpora. The task is highly practical since non-parallel multilingual data exist in far greater quantities than parallel corpora, but parallel sentences are a much more useful resource. Here we propose a web crawling method for building subject-aligned comparable corpora from Wikipedia articles. We also introduce a method for extracting truly parallel sentences that are filtered out from noisy or just comparable sentence pairs. We describe our implementation of a specialized tool for this task as well as training and adaption of a machine translation system that supplies our filter with additional information about the similarity of comparable sentence pairs.

## 1  Introduction

Parallel sentences are an invaluable information resource especially for machine translation systems as well as for other cross lingual information dependent tasks. Unfortunately such data is quite rare, especially for the Polish–English language pair. On the other hand, monolingual data for those languages is accessible in far greater quantities. We can classify the similarity of data as four main corpora types. Most rare parallel corpora can be defined as corpora that contain translation of the same document into two or more languages. Such data should be aligned at least at the sentence level. A noisy-parallel corpus contains bilingual sentences that are not perfectly aligned or has poor quality translations. Nevertheless mostly bilingual translation of a specific document should be present in it. A comparable corpus is built from non-sentence-aligned and not-translated bilingual documents, but the documents should be topic-aligned. A quasi-comparable corpus includes very heterogeneous and very non-parallel bilingual documents that can but don't have to be topic-aligned.

In this article we present a methodology that allows us to obtain truly parallel corpora from not sentence-aligned data sources, such as noisy-parallel or comparable corpora. For this purpose we used a set of specialized tools for obtaining, aligning, extracting and filtering text data, combined together into a pipeline that allows us to complete the task. We present the results of our initial experiments based on randomly selected text samples from Wikipedia. We chose Wikipedia as a source of data because of a large number of documents that it provides (1,047,423 articles on PL Wiki and 4,524,017 on EN, at the time of writing this article). Furthermore, Wikipedia contains not only comparable documents, but also some documents that are translations of each other. The quality of our approach is compared to human evaluation.

The solution can be divided into three main steps. First the data is collected, then it is aligned, and lastly the results of the alignment are filtered. The last two steps are not trivial because of the disparities between Wikipedia documents. Based on the Wikipedia statistics we know that an average article on PL Wiki contains about 379 words, whereas on EN Wiki it is 590 words. This is most likely why sentences in the raw Wiki corpus are mostly misaligned, with translation lines whose placement does not correspond to any text lines in the source language. Moreover, some sentences may have no corresponding translation in the corpus at all. The corpus might also contain poor or indirect translations, making the alignment difficult. Thus, alignment is crucial for accuracy. Sentence alignment must also be computationally feasible in order to be of practical use in various applications.

The Polish language presents a particular challenge to the application of such tools. It is a complicated West-Slavic language with complex elements and grammatical rules. In addition, the Polish language has a large vocabulary due to many endings and prefixes changed by word declension. These characteristics have a significant impact on the data and data structure requirements.

In addition, English is a position-sensitive language. The syntactic order (the order of words in a sentence) plays a very significant role, and the language has very limited inflection of words (due to the lack of declension endings). The word position in an English sentence is often the only indicator of the meaning. The sentence order follows the Subject-Verb-Object (SVO) schema, with the subject phrase preceding the predicate. On the other hand, no specific word order is imposed in Polish, and the word order has little effect on the meaning of a sentence. The same thought can be expressed in several ways. For example, the sentence "I bought myself a new car." can be written in Polish as one of the following: "Kupiłem sobie nowy samochód"; "Nowy samochód sobie kupiłem."; "Sobie kupiłem nowy samochód."; "Samochód nowy sobie kupiłem.". It must be noted that such differences exist in many language pairs and need to be dealt with in some way.

## 2 The pipeline

Our procedure starts with a specialized web crawler. Because PL Wiki contains less data of which almost all articles have their correspondence on EN Wiki the program crawlers data starting from non-English site first. It is a language independent solution. The crawler can obtain and save bilingual articles of any language supported by Wikipedia.

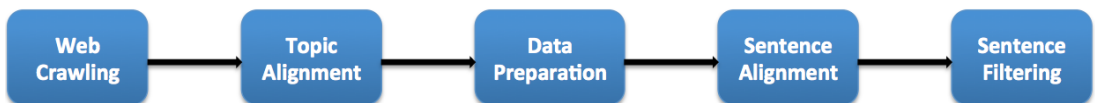

**Figure 1. Pipeline**

First the data is saved in HTML files and then it is topic-aligned. In order to narrow the search field to a specific in-domain documents, it is necessary to give the crawler the first link to the article in domain, and then the program will automatically obtain other topic related documents. Narrowing the search domain not only helps to adjust the output to the specific needs, but also narrows the vocabulary, which makes the aligning task easier. After obtaining HTML documents, the crawler extracts plain text from them and cleans the data. Tables, URL's, figures, pictures, menus, references and other unnecessary data are removed. Finally, bilingual documents are tagged with a unique ID as a topic-aligned comparable corpus.

We propose a two-level sentence alignment method that prepares a dictionary for itself. The Hunalign tool is used first to match bilingual sentences. Its input is tokenized and sentence-segmented. In the presence of a dictionary, Hunalign combines the dictionary information with the Gale-Church sentence-length information. In the absence of a dictionary, it first falls back to the sentence-length information, and then builds an automatic dictionary based on this alignment. Then it realigns the text in a second pass, using the automatic dictionary. The option without a dictionary is the one we used.

Like most sentence aligners, Hunalign does not deal well with changes of the sentence order. It is unable to come up with crossing alignments, i.e., segments A and B in one language corresponding to segments B' A' in the other language. In order to cope with this problem and filter out bad or poor bilingual sentence pairs, we implemented a special tool.

## 2.1 Filtering strategy

Our strategy is to find a correct translation of each Polish line using any translation engine. We translate all lines of the Polish file (src.pl) with a translator and put each line translation in an intermediate English translation file (src.trans). This intermediate translation helps us find the correct line in the English translation file (src.en) and put it in the correct position or remove incorrect pairs from the corpora. There are additional complexities that must be addressed. Comparing the src.trans lines with the src.en lines is not easy, and it becomes harder when we want to use the similarity rate to choose the correct, real-world translation.

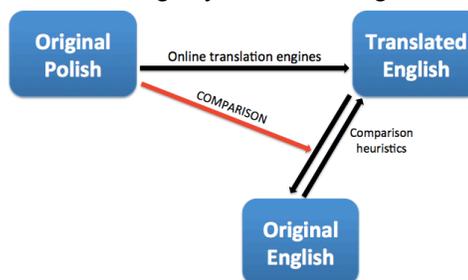

**Figure 2. Filtering**

There are many strategies to compare two sentences. We can split each sentence into its words and find the number of words in both sentences. However, this approach has some problems. For example, let us compare "It is origami." to these sentences: "The common theme what makes it origami is folding is how we create the form."; "This is origami."

With this strategy, the first sentence is more similar because it contains all 3 words. However, it is clear that the second sentence is the correct choice. We can solve this problem by dividing the number of words in both sentences by the number of total words in the sentences. However, counting stop words in the intersection of sentences sometimes causes incorrect results. So, we remove these words before comparing two sentences.

Another problem is that sometimes we find stemmed words in sentences, for example "boy" and "boys." Despite the fact that these two words should be counted as similarity of two sentences, with this strategy, these words are not counted.

The next comparison problem is the word order in sentences. There are other ways for comparing strings that are better than counting intersection lengths. For example, we can find matching blocks in the strings "abxcd" and "abcd".

Our function can count "ratio" and divide the length of matching blocks by the length of two strings, and return a measure of the sequences' similarity as a float value in the range [0, 1]. This measure is $2.0*M / T$, where T is the total number of elements in both sequences, and M is the number of matches. Using this function to compare strings instead of counting similar words helps us solve the problem of the similarity of "boy" and "boys". It also solves the problem of considering the position of words in sentences.

Another problem in comparing lines is synonyms. For example, these two sentences: "I will call you tomorrow."; "I would call you tomorrow.". We used the NLTK Python module and WordNet® to find synonyms for each word and to use these synonyms in comparing sentences. Using synonyms for each word, we created multiple sentences from each original sentence and compared them as a many-to-many relation.

To obtain the best results, our script provides users with the ability to have multiple functions with multiple acceptance rates. Fast functions with lower quality results are tested first. If they can find results with a very high acceptance rate, we accept their selection. If the acceptance rate is not sufficient, we use slower but higher accuracy functions.

## 2.2 Wikipedia Machine Translation Engine

The filtering tool, which is most important part in entire process, is dependent on the translation engine. It is possible to use online engines for general use, but better results can be obtained with specialized translation systems. We obtained all PL-EN parallel data from various domains from the OPUS project and used it for training a specialized machine translation system. To improve its performance, we conducted the system's adaptation to Wikipedia using a dump of all English articles as a language model. The final training corpora counted 36,751,049 sentences and the language model counted 79,424,211 sentences. The unique word forms count was 3,209,295 in the Polish side of the corpora, 1,991,418 in the English side and 37,702,319 in the language model. Implementation of the translation system included many steps. Processing of the corpora was accomplished, including tokenization, cleaning, factorization, lowercasing, splitting, and a final cleaning after splitting. Training data was processed and the language model was developed. Tuning was performed as well.

The training was done using the Moses open source SMT toolkit with its Experiment Management System (EMS). The SRI Language Modeling Toolkit (SRILM) with an interpolated version of the Kneser-Key discounting (interpolate –unk –kndiscount) was used for the 6-gram language model training. We used the MGIZA++ tool for word and phrase alignment. KenLM was used to binarize the language model, with a lexical reordering set to use the msd-bidirectional-fe model. Reordering probabilities of phrases were conditioned on lexical values of a phrase. It considers three different orientation types on source and target phrases like monotone(M), swap(S) and discontinuous(D). The bidirectional reordering model adds probabilities of possible mutual positions of source counterparts to the current and following phrases. Probability distribution to a foreign phrase is determined by "f" and to the English phrase by "e". MGIZA++ is a multi-threaded version of the well-known GIZA++ tool. The symmetrization method was set to grow-diag-final-and for word alignment processing. First, two-way direction alignments obtained from GIZA++ were intersected, so only the alignment points that occurred in both alignments remained. In the second phase, additional alignment points existing in their union were added. The growing step adds potential alignment points of unaligned words and neighbours. Neighbourhood can be set directly to left, right, top or bottom, as well as to diagonal (grow-diag). In the final step, alignment points between words from which at least one is unaligned are added (grow-diag-final). If the grow-diag-final-and method is used, an alignment point between two unaligned words appears.

### 2.2.1 MT Evaluation

Metrics are necessary to measure the quality of translations produced by the SMT systems. For this purpose, various automated metrics are available to compare SMT translations to high quality human translations. Since each human translator produces a translation with different word choices and orders, the best metrics measure SMT output against multiple reference human translations. Among the commonly used SMT metrics are: Bilingual Evaluation Understudy (BLEU), the U.S. National Institute of Standards & Technology (NIST) metric, the Metric for Evaluation of Translation with Explicit Ordering (METEOR), Translation Error Rate (TER).

BLEU was one of the first metrics to demonstrate a high correlation with reference human translations. The general approach for BLEU, as described in [do SMT z madery], is to attempt to match variable length phrases to reference translations. Weighted averages of the matches are then used to calculate the metric.

The NIST metric seeks to improve the BLEU metric by valuing information content in several ways. It takes the arithmetic versus geometric mean of the *n*-gram matches to reward good translation of rare words. The NIST metric also gives heavier weights to rare words. Lastly, it reduces the brevity penalty when there is a smaller variation in the translation length.

The METEOR metric, developed by the Language Technologies Institute of Carnegie Mellon University, is also intended to improve the BLEU metric. We used it without synonym and paraphrase matches for Polish. METEOR rewards recall by modifying the BLEU brevity penalty, takes into account higher order *n*-grams to reward matches in a word order, and uses arithmetic vice geometric averaging. For multiple reference translations, it reports the best score for word-to-word matches.

TER is one of the most recent and intuitive SMT metrics developed. This metric determines the minimum number of human edits required for an SMT translation to match a reference translation in meaning and fluency. Required human edits might include inserting words, deleting words, substituting words, and changing the order or words or phrases.

For the evaluation, we randomly selected 1000 parallel sentences from Wikipedia documents. None of those sentences were included inside the training data on our system. Table 1 presents evaluation of translation quality in comparison to general use online translation engines.

|  | BLEU | NIST | METEOR | TER |
|---|---|---|---|---|
| Google | 18,15 | 5,22 | 48,86 | 70,23 |
| Bing | 18,87 | 5,27 | 48,80 | 70,61 |
| Our SMT | 20,51 | 5,31 | 49,23 | 69,11 |

**Table 1. MT Results**

# 3   Experiments and mining evaluation

To evaluate quality and quality of parallel data, extracted automatically from comparable corpora, we randomly selected 20 bilingual documents from Wikipedia. Some of them differed greatly in the means with respect to vocabulary, text amounts and parallelism. We asked human translators to manually align those articles on the sentence level. The information about the human translators is presented in Table 2.

|  | Vocab.Count | | Sentences | | Human Aligned | |  | Vocab.Count | | Sentences | | Human Aligned |
|---|---|---|---|---|---|---|---|---|---|---|---|---|
|  | PL | EN | PL | EN | YES | NO |  | PL | EN | PL | EN | YES |
| 1 | 2526 | 1910 | 324 | 186 | 127 |  | 11 | 2861 | 2064 | 412 | 207 | 8 |
| 2 | 2950 | 3664 | 417 | 596 | 6 |  | 12 | 2186 | 1652 | 345 | 188 | 2 |
| 3 | 504 | 439 | 45 | 43 | 34 |  | 13 | 2799 | 3418 | 496 | 472 | 124 |
| 4 | 2529 | 1383 | 352 | 218 | 4 |  | 14 | 1164 | 1037 | 184 | 196 | 3 |
| 5 | 807 | 1666 | 104 | 275 | 10 |  | 15 | 2465 | 1781 | 365 | 189 | 3 |
| 6 | 2461 | 4667 | 368 | 618 | 1 |  | 16 | 1946 | 1839 | 282 | 198 | 132 |
| 7 | 2701 | 1374 | 560 | 210 | 16 |  | 17 | 966 | 782 | 113 | 96 | 7 |
| 8 | 1647 | 768 | 274 | 78 | 1 |  | 18 | 2005 | 1253 | 309 | 134 | 1 |
| 9 | 1189 | 1247 | 121 | 120 | 64 |  | 19 | 2443 | 2001 | 251 | 189 | 21 |
| 10 | 2933 | 3047 | 296 | 400 | 150 |  | 20 | 9888 | 3181 | 1297 | 367 | 2 |

**Table 2. Human alignment**

The same articles were processed with our pipeline. In Table 3 we present how many sentences Hunalign initially aligned as similar and how many of them remained after filtering with our tool. Both columns "YES" and "NO" under the Hunaligned section are aligned sentences, the number represent how many of them were aligned correctly and how many by mistake. In Filtered column we present number parallel sentences remained after filtering, in "YES" column we show properly aligned sentences and "NO" mistaken ones. In this scenario, we also asked a human translator to check which of the remaining sentence pairs were truly parallel and if any pairs were missed out.

|    | Hunaligned | | Filtered | | wywalic | |    | Hualigned | | Filtered | | wywalic | |
|----|-----|-----|-----|----|-----|---|----|-----|------|-----|----|-----------|---|
|    | YES | NO  | YES | NO | No. | % |    | YES | NO   | Yes | No | Diff. No. | % |
| 1  | 109 | 130 | 18  | 0  |     |   | 11 | 8   | 325  | 0   | 0  |           |   |
| 2  | 6   | 527 | 25  | 2  |     |   | 12 | 2   | 256  | 0   | 0  |           |   |
| 3  | 17  | 24  | 0   | 0  |     |   | 13 | 70  | 414  | 1   | 0  |           |   |
| 4  | 4   | 302 | 1   | 0  |     |   | 14 | 3   | 182  | 0   | 0  |           |   |
| 5  | 6   | 211 | 1   | 0  |     |   | 15 | 3   | 285  | 0   | 0  |           |   |
| 6  | 1   | 498 | 0   | 0  |     |   | 16 | 111 | 108  | 0   | 0  |           |   |
| 7  | 16  | 440 | 0   | 0  |     |   | 17 | 7   | 98   | 0   | 0  |           |   |
| 8  | 1   | 221 | 0   | 0  |     |   | 18 | 1   | 202  | 0   | 0  |           |   |
| 9  | 51  | 62  | 0   | 0  |     |   | 19 | 21  | 192  | 0   | 0  |           |   |
| 10 | 127 | 245 | 0   | 0  |     |   | 20 | 2   | 1078 | 0   | 0  |           |   |

**Table 3. Automatic alignment**

# 4 Conclusions and future work

We introduced a new method for obtaining, mining and filtering very parallel bilingual sentence pairs from noisy-parallel and comparable corpora. Nowadays, the bi-sentence extraction task is becoming more and more popular in unsupervised learning for numerous specific tasks.   The method overcomes disparities between English and Polish or any other West-Slavic languages. It is a language independent method that can easily be adjusted to a new environment, and it only requires parallel corpora for initial training. The experiments show that the method provides good accuracy and some correlation with human evaluations. That is what should be expected from the task of mining from comparable data. From a practical standpoint, the method neither requires expensive training nor requires language-specific grammatical resources, while producing results with good accuracy.

Nevertheless, there is still some room for improvement in two areas. In presented experiments the amount of obtained data in comparison with human work is not satisfactory. The first one is Hunalign, which would perform much better if it was provided a good quality dictionary, especially such that contains in-domain vocabulary. The second one is the statistical machine translation system (SMT), which would greatly increase quality by providing better translations. After the initial mining of the corpora, the obtained parallel data can possibly be used for both purposes. Firstly, a phrase-table can be trained from extracted bi-sentences and from it we can easily extract a good in-domain dictionary (also including probabilities of translations). Secondly, the SMT can be retrained with newly mined data and adapted based on it. Lastly, the pipeline can be re-run with new capabilities. The steps can be repeated until the extraction results are satisfactory.

# 5 Acknowledgements

This work was supported by the European Community from the European Social Fund within the Interkadra project UDA-POKL-04.01.01-00-014/10-00 and Eu-Bridge 7th FR EU project (Grant Agreement No. 287658).